\documentclass[conference]{IEEEtran}
\IEEEoverridecommandlockouts
\usepackage{cite}
\usepackage{amsmath,amssymb,amsfonts}
\usepackage{algorithmic}
\usepackage{graphicx}
\usepackage{textcomp}
\usepackage{xcolor}
\usepackage{multirow}
\usepackage{booktabs}
\usepackage{array}
\usepackage{graphicx} % For \resizebox
\usepackage{amsmath}
\usepackage{subcaption}
\usepackage{enumitem}
\usepackage{hyperref}
\usepackage{cleveref}
\usepackage{xurl}
\begin{document}
\title{EF-Net: A Deep Learning Approach Combining Word Embeddings and Feature Fusion for Patient Disposition Analysis}

\author{\IEEEauthorblockN{Nafisa Binte Feroz$^{1}$, Chandrima Sarker$^{1}$, Tanzima Ahsan$^{1}$, K M Arefeen Sultan$^{2}$, Raqeebir Rab$^{1}$}
\IEEEauthorblockA{$^{1}$Dept of Computer Science and Engineering,\\ Ahsanullah University of Science \& Technology, Dhaka, Bangladesh}
\IEEEauthorblockA{$^{2}$Scientific Computing and Imaging Institute,\\University of Utah, Utah, USA}
\IEEEauthorblockA{Email: nbinteferoz67@gmail.com,   shipra.rpsc@gmail.com, tanzimaahsan431@gmail.com,\\
arefeen.sultan@utah.edu, raqeebir.cse@aust.edu}
%\and

}
\maketitle

\begin{abstract}
One of the most urgent problems is the overcrowding in emergency departments (EDs), caused by an aging population and rising healthcare costs. Patient dispositions have become more complex as a result of the strain on hospital infrastructure and the scarcity of medical resources. Individuals with more dangerous health issues should be prioritized in the emergency room. Thus, our research aims to develop a prediction model for patient disposition using EF-Net. This model will incorporate categorical features into the neural network layer and add numerical features with the embedded categorical features. We combine the EF-Net and XGBoost models to attain higher accuracy in our results. The result is generated using the soft voting technique. In EF-Net, we attained an accuracy of 95.33\%, whereas in the Ensemble Model, we achieved an accuracy of 96\%. The experiment's analysis shows that EF-Net surpasses existing works in accuracy, AUROC, and F1-Score on the MIMIC-IV-ED dataset, demonstrating its potential as a scalable solution for patient disposition assessment. Our code is available at https://github.com/nafisa67/thesis
\end{abstract}


\section{Introduction}
The global demand for emergency department (ED) services is increasing, resulting in overcrowding and a subsequent decrease in the quality of care\cite{feretzakis2024machine}. An increase in demand has been linked to higher morbidity and mortality rates, which are aggravated by care delays. Hospitals must effectively manage their resources, including the availability of medical equipment, personnel, and beds. Identifying patients in a deteriorating condition at an early stage is essential, as the prognosis of critical patients is relatively susceptible to opportune interventions\cite{chen2024using}. Emergency physicians consistently face the challenging task of identifying patients who require hospitalization and providing them with the necessary medical care. Medical professionals base their decisions on experience and test results, but occasionally personal issues cloud their judgment and increase the risk of illness or even death. Consequently, the development of a clinical decision support system that predicts the disposition of emergency patients can contribute to the optimization of the allocation of medical resources, the reduction of human error, the enhancement of the quality of medical care, and importance will be given to the severe patient\cite{hautz2019diagnostic}.\\
Many studies are presently being conducted to construct prediction models that use machine learning techniques to anticipate patient attitudes in hospital emergency departments.\cite{chen2024using} \cite{feretzakis2024machine}. The predictive models in \cite{kuo2024ensemble} employ a variety of data types, such as structured information like demographic details and vital signs, unstructured information like triage notes and chief complaints, or a combination of both structured and unstructured data. While these studies have greatly advanced our understanding of ED dispositions, the majority of the models they create are limited to predicting one binary disposition at a time, such as discharge versus admission, with the exception of \cite{kuo2024ensemble}. This might not always be possible when there are more than two possible ED dispositions. Furthermore, not all of ensemble learning's potential has been fully explored. Combining several distinct classifiers into an ensemble learning technique allows for better classification performance than would be possible with each one working alone\cite{kuo2024ensemble}.

In this paper, we aim to develop a deep learning-based fusion model of word embeddings and numerical features named EF-Net, where we utilize both the categorical and numerical feature representation in latent space for multiclass forecasting of patient dispositions in the emergency department. Additionally, we developed an ensemble learning-based model with EF-Net and XGBoost that predicts multiple disposition outcomes for ED patients simultaneously. Experimental findings reveal that EF-Net surpasses previous works in achieving higher accuracy, AUROC, Precision, Recall, and F1-Score. The contributions of this paper are summarized as follows.
\begin{itemize}
    \item Introducing word embedding for categorical features and performing a fusion of both numerical and categorical features on latent space. This helps the network to learn more meaningful semantic features in the lower feature space representation thus boosting the performance in the output layer.
    \item We also perform an ensemble of both our model EF-Net and XGBoost, showing with experiments that our model is capable of being incorporated in ensemble models.
    \item Extensive experiments and comparisons with previous methods demonstrate the generality, effectiveness, and efficiency of our model.
\end{itemize}
%in these studies, with the exception of \cite{kuo2024ensemble} \cite{arnaud2020deep} \cite{klang2021predicting} \cite{kuo2024ensemble} \cite{klang2021simple}. 

%\section{Problem statement}

\section{Related Works}
Some recent studies have aimed to create predictive models for emergency department dispositions. Most of the studies evaluate their model using MIMIC-IV-ED dataset \cite{mimic-iv-ed}. Using the MIMIC-IV-ED dataset\cite{mimic-iv-ed}, Kuo \emph{et al.} \cite{kuo2024ensemble} concentrated on applying machine learning, more especially ensemble learning, to predict patient dispositions in emergency departments (ED). They used structured data (demographics, vital signs) and unstructured data (chief complaints, preliminary diagnoses). They employed Bag-of-Words (BOW) and Term Frequency-Inverse Document Frequency (TF-IDF) to preprocess the unstructured data. \cite{kuo2024ensemble}  used simple base learners such as Random Forest and a Multi-Layer Perceptron (MLP) as the meta-learner.Feretzakis \emph{et al.} \cite{feretzakis2024machine}  emphasized the importance of efficient triage and introduced an automated machine-learning (AutoML) approach using the MIMIC-IV-ED dataset\cite{mimic-iv-ed} to predict patient dispositions at triage. It primarily relies on structured data, such as vital signs, and does not incorporate unstructured data, such as triage notes.

%The inclusion of triage notes and other textual data allows the model to capture more complex patterns, which are crucial in emergency settings.  \cite{feretzakis2024machine} do not utilize ensemble learning techniques, which could enhance model robustness and performance. We incorporate ensemble learning by combining EF-Net with XGBoost, resulting in higher accuracy and better overall performance. They focus on a limited set of data, primarily initial vital signs and basic metrics. But we integrate a broader range of data types, including unstructured data, to enhance predictive capabilities.

Sulaiman \emph{et al.} \cite{sulaiman2023emergency} explored the use of explainable AI (XAI) in predicting hospitalizations during Emergency Department (ED) triage using the MIMIC-IV-ED dataset and Gradient Boosting (GB) modeling. Assaf \emph{et al.} \cite{assaf202030} aim to develop a predictive model for 30-day readmission using the MIMIC III dataset. For each visit, features are extracted from the dataset's vital signs and ICD codes. Li \emph{et al.} \cite{li2020hospital} addressed the problem of class imbalance in ICU in-hospital mortality prediction by using a balanced random forest (BRF) algorithm. They employed limited features for prediction, including vital signs, ICD codes, and gender. Although ICD code embeddings are used, other potentially valuable data types, such as patient notes, medication details, and more extensive demographic information, are not included. They also introduced a novel performance metric, modified arithmetic mean (AGm), for effective model selection. 

%Where as, we incorporate a broader range of data types, including both structured (e.g., vital signs, demographics) and unstructured data (e.g., triage notes)

%In EF-Net, we integrate a broader range of data types, including both structured (e.g., vital signs, demographics) and unstructured data (e.g., triage notes).
Gentimis \emph{et al.} \cite{gentimis2017predicting} address the absence of predictive models for the length of hospital stay independent of specific diseases. Two key algorithms, Neural Networks and Random Forests, are employed for analysis. %Random Forests, an ensemble method, combines decision trees for robust modeling.
Melissourgos \emph{et al.} \cite{melissourgos2021outsourcing} propose a model that used matrix masking to protect patient data privacy when outsourcing medical data to the cloud for training artificial neural networks (ANNs). In another study, Dhanalakshmi \emph{et al.} \cite{dhanalakshmi2020predicting} focused on feature extraction's importance in predicting re-admission. These features include free text clinical notes for diagnosis, key phrase extraction, 30-day re-admission prediction, displaying these results in an online application, and predicting the likelihood of re-admitting due to the coronavirus.

Most of the approaches primarily focus on structured data for disposition classification. However, in \cite{kuo2024ensemble}, \cite{feretzakis2024machine}, the management of unstructured data is relatively basic, concentrating on word frequency without deeper semantic understanding.
%Most of the approaches focused only on structured data for binary classification. They did not handle unstructured data.\cite{assaf202030} \cite{feretzakis2024machine} \cite{li2020hospital}. Tsoni \cite{tsoni2023machine} \emph{et al.} used structured data from the MIMIC-IV-ED dataset and data was preprocessed using KNIME. However, handling unstructured data is relatively basic, focusing on word frequency without deeper semantic understanding. 
Our EF-Net model integrates structured and unstructured data more effectively by leveraging word embedding, which captures the semantic meaning of the text. This approach goes beyond simple word counts and can understand the context in which words appear, leading to more accurate predictions. Our new model combines word embeddings and feature fusion, enabling it to capture more intricate patterns in the data.

\section{Dataset}
\begin{figure}[ht]
    \centering
    \includegraphics[width=1\linewidth]{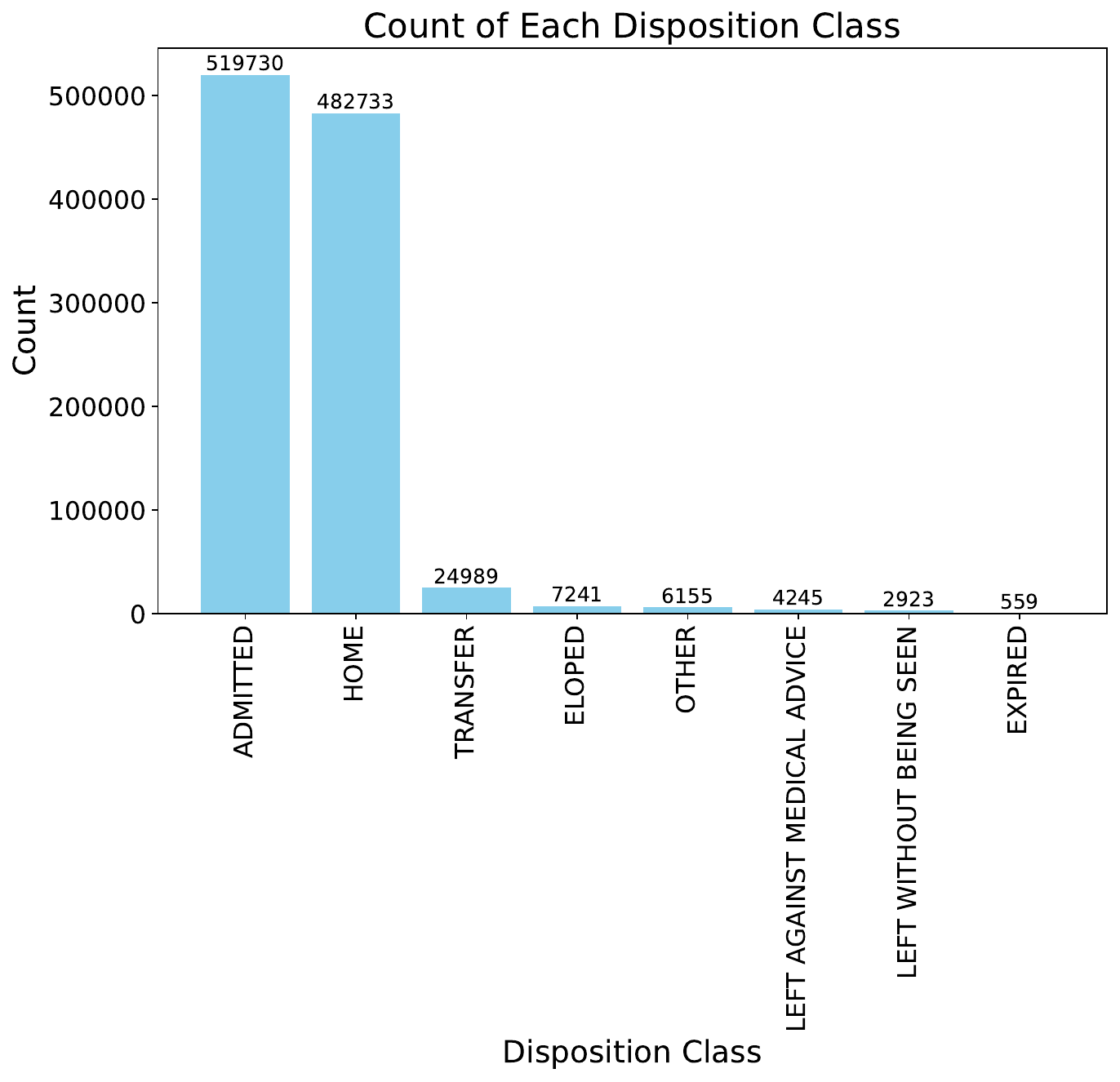}
    \caption{Distribution of Disposition Classes in MIMIC-IV ED dataset.}
    \label{fig:disposition_dist}
\end{figure}

The data used in this investigation were obtained from the Medical Information Mart for Intensive Care MIMIC-IV-ED v2.0 dataset\cite{mimic-iv-ed}. \cite{mimic-iv-ed} presents data regarding patient admissions to the emergency department (ED) at Beth Israel Deaconess Medical Center from 2011 to 2019. The dataset comprises around 425,000 occurrences of emergency department admissions. The MIMIC-IV-ED v2.0 dataset encompasses data on vital signs, triage details, drug administration, medication reconciliation, and discharge diagnoses for 132,197 patients. 
%There are available vital signs, triage details, medication delivery, medication reconciliation, and discharge diagnoses. 
 %The goal of MIMIC-IV-ED is to provide assistance for a wide range of research projects and educational initiatives. Patient identifiers were removed according to the Health Insurance Portability and Accountability Act (HIPAA) Safe Harbor provision. 
The dataset originally has 1,048,575 rows and 26 columns. 14 are numerical, and 12 are categorical data. The numerical columns include subject\_id, hadm\_id, stay\_id, temperature, heart rate, respiratory rate, oxygen saturation, systolic blood pressure, diastolic blood pressure, ICD version, GSN, NDC, etccode, acuity, etc. The categorical columns include gender, race, arrival transport, chart time, rhythm, pain, chiefcomplaint, icd\_code, icd\_title, name, etcdescription, disposition. MIMIC-IV-ED v2.0 comprises eight disposition classes such as admission, home, transfer, eloped, left against medical advice, left without being seen, expired and other; Figure \ref{fig:disposition_dist} depicts the distribution of these classes. The bar plot illustrates the counts of each disposition classes.

% \pagebreak
\section{Methodology}
\begin{figure}[ht]
    \centering
    \includegraphics[width=0.8\linewidth]{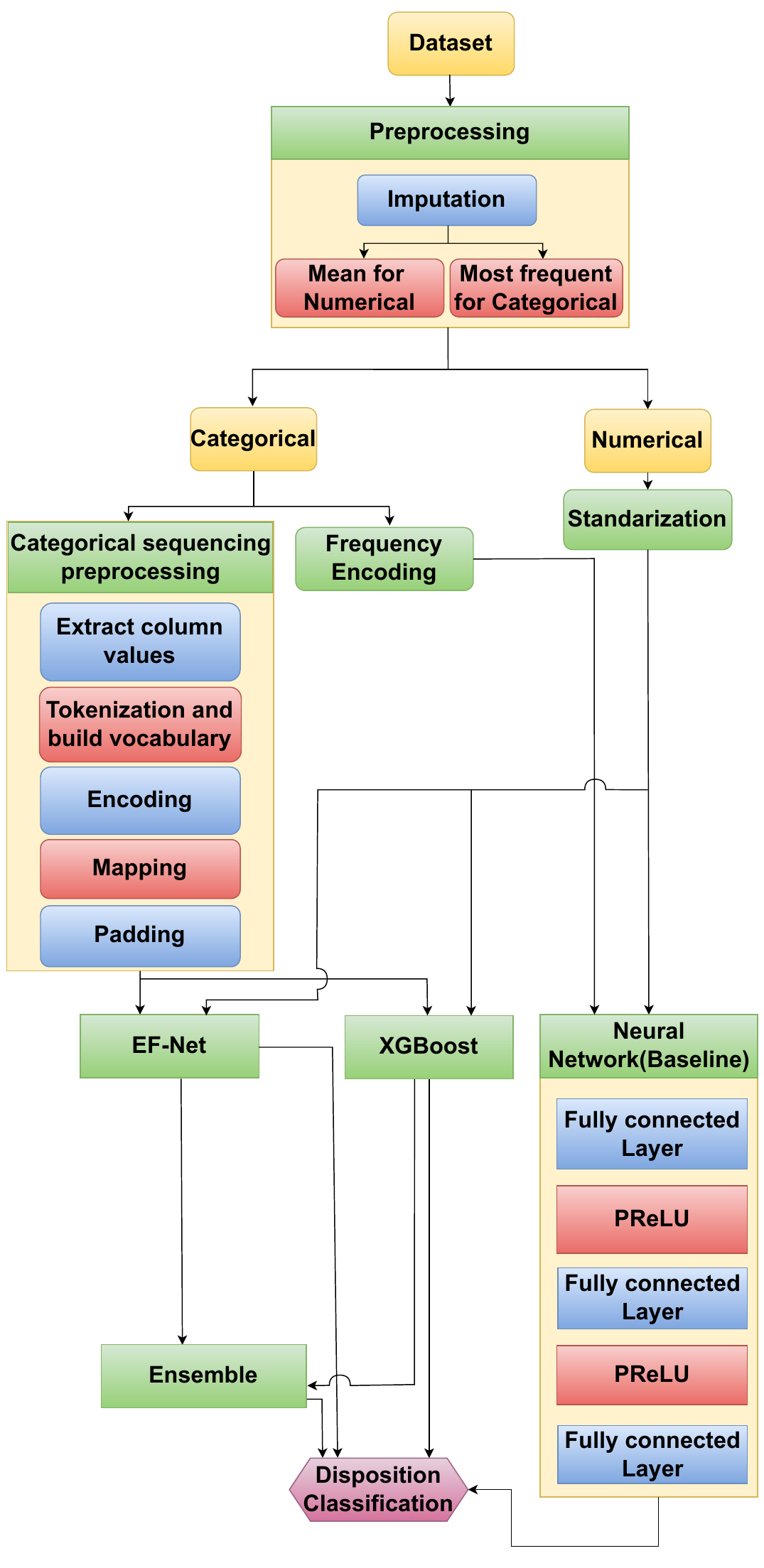}
    \caption{Flowchart of Proposed Methodology.}
    \label{fig:flow-chart}
\end{figure}

This section presents our proposed methodology for disposition classification in the Emergency Department (ED).  Depicted in Figure\ref{fig:flow-chart} at first, we discuss about the data preprocessing. Afterwards, we formulate our baseline neural network model. Subsequently, we discuss regarding our proposed methodology built around a baseline neural network. Finally, we talk about the Ensemble of XGBoost and EF-Net. Figure \ref{fig:flow-chart} depicts functional implementations for forecasting the patient's disposition in the ED. 

\subsection{Data Preprocessing}
We began by identifying the numerical and categorical columns. Numerical features are represented as \( \mathbf{N} \in \mathbb{R}^{B \times N} \), where \( B \) denotes the batch size and \( N \) is the number of numerical features, which is 14. Categorical features were denoted as \( \mathbf{C} \in \mathbb{R}^{B \times C} \), where \( C \) is the number of categorical features, which is 12. During preprocessing, missing numerical values were imputed with the mean:
\[
\mathbf{N}_{\text{imputed}} = \mathbf{N}_{\text{missing}} \text{ replaced with } \mu(\mathbf{N}),
\]
where \( \mu(\mathbf{N}) \) represents the mean. Missing categorical values were imputed with the mode:
\[
\mathbf{C}_{\text{imputed}} = \mathbf{C}_{\text{missing}} \text{ replaced with } \text{mode}(\mathbf{C}).
\]
The numerical features were standardized using:
\[
\mathbf{N}_{\text{scaled}} = \frac{\mathbf{N}_{\text{imputed}} - \mu(\mathbf{N})}{\sigma(\mathbf{N})},
\]
where \( \sigma(\mathbf{N}) \) is the standard deviation. Categorical features were tokenized by converting them to lowercase:
\[
\mathbf{C}_{\text{tokenized}} = \text{Lower}(\mathbf{C}_{\text{imputed}}).
\]
Each column's vocabulary was mapped to unique indices, and the encoded values were padded:
\[
\mathbf{C}_{\text{padded}} = \text{Pad}(\mathbf{C}_{\text{encoded}}),
\]
ensuring consistent length \( L \) across all columns. The padded data was concatenated, resulting in \( S \) columns, where \( S = \sum_{j=1}^{C} L_j \) is the total length across all categorical features. The column grew to 77, which is the sum of all padded data columns' maximum length.

\textbf{Handling the class imbalance problem} 
To address the class imbalance problem, we applied stratified splitting:
\[
(X_{\text{train}}, X_{\text{val}}, X_{\text{test}}, y_{\text{train}}, y_{\text{val}}, y_{\text{test}}) = \text{StratifiedSplit}(X, y).
\]
where $X\_{\text{train}}$, $y\_{\text{train}}$ represent the features for train the model, $X\_{\text{val}}$,$y\_{\text{val}}$ means the the features for validation, $X\_{\text{test}}$, $y\_{\text{test}}$ represents the features for test the model.
The stratified division ensures that each class is represented proportionally in the training, validation, and test datasets despite the possibility of biased models arising from class imbalances in datasets.

\subsection{Baseline Model for Predicting ED Desposition}
Initially, we created a neural network model as our baseline model for forecasting ED outcomes using all the features. We made a neural network for disposition classification that has three fully connected layers and uses PReLU after the first two layers for nonlinearity. The neural network processed the standardized numerical features and frequency encoded categorical features.

\subsection{Proposed EF-Net Model}
\begin{figure}[ht]
    \centering
    \includegraphics[width=1\linewidth]{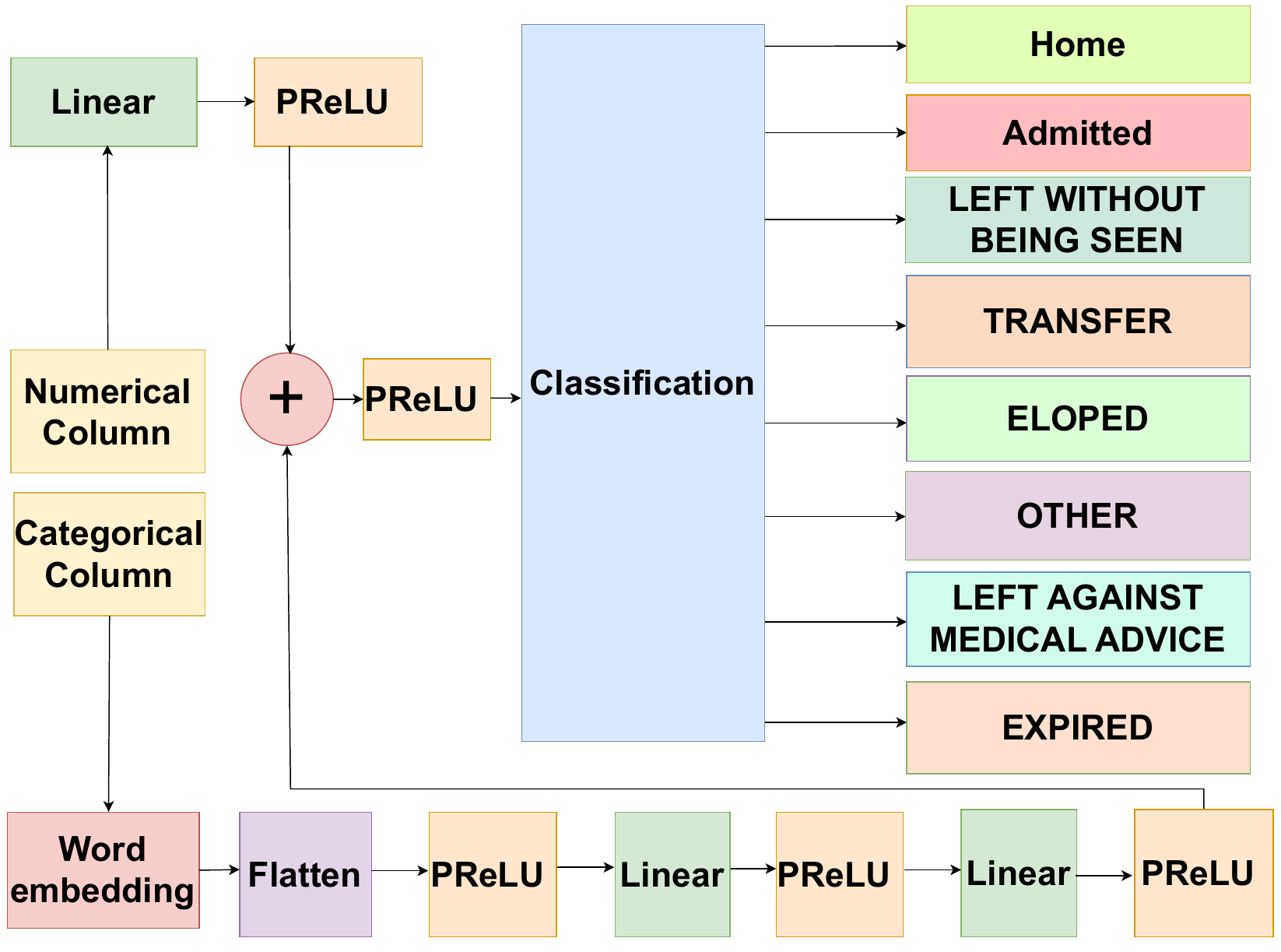}
    \caption{Architecture of EF-Net Model.}
    \label{fig:efnet_arch}
\end{figure}

In our proposed EF-Net model, we employ a combination of word embedding, linear transformations, and activation functions to effectively process and classify the input data. The architecture consists of several key layers, each contributing to the model's overall performance, as outlined in \Cref{tab:model_architecture}. The overall architecture of EF-Net model is shown in Fig \ref{fig:efnet_arch}. Our model processes numerical/categorical columns separately, followed by feature fusion. Numerical inputs are passed through a linear layer and PReLU activation, while word embeddings of categorical column undergo Flattening, Linear layers, and PReLU. The fused features are processed through a PReLU layer and classified into various outcomes such as "Home," "Admitted," "Transfer," and others. The working architecture of EF-net model is given below:

\begin{itemize}
    \item The model begins with a \textbf{Word Embedding} layer, where categorical input features \(c \in \mathbb{R}^{B \times C}\) are transformed into dense vector representations \(E(c) \in \mathbb{R}^{B \times C \times d}\). This embedding step captures the semantic relationships between categorical features, which are essential for downstream tasks.

    \item Next, the embedded features are \textbf{Flattened}, resulting in a vector \(F(E(c)) \in \mathbb{R}^{B \times (C \times d)}\). This step prepares the data for subsequent linear operations by reducing the dimensionality of the embedding space.

    \item Following the flattening, the model applies a \textbf{Linear} transformation to the flattened embeddings, producing an output \(L_1(F(E(c))) \in \mathbb{R}^{B \times 32}\). This is followed by a PReLU activation, which introduces non-linearity and enhances the model's ability to capture complex patterns in the data.

    \item The processed embeddings are further refined through another \textbf{Linear} layer, which maps the output to a lower-dimensional space \(L_2(L_1) \in \mathbb{R}^{B \times 16}\), followed by another PReLU activation.

    \item Simultaneously, the model processes the numerical features \(n \in \mathbb{R}^{B \times N}\) through a separate \textbf{Linear} layer, producing an output \(L_n(n) \in \mathbb{R}^{B \times 16}\), which is also activated using PReLU. This step aligns the dimensionality of numerical features with that of the processed categorical features, allowing for their subsequent integration.

    \item The integration of categorical and numerical features is achieved through an \textbf{Addition} operation, where the outputs \(L_2\) and \(L_n\) are summed, resulting in a combined representation \(S \in \mathbb{R}^{B \times 16}\). This summation allows the model to leverage information from both feature types simultaneously.

    \item Finally, the combined features are passed through a \textbf{Linear} classification layer, which outputs the predictions \(y \in \mathbb{R}^{B \times K}\). This final layer maps the processed data to the desired number of output classes, completing the model's forward pass.
\end{itemize}

\begin{table*}[t]
    \centering
    \resizebox{0.8\linewidth}{!}
    {\begin{tabular}{|l|c|c|l|}
        \hline
        \textbf{Layer Type} & \textbf{Input} & \textbf{Output} & \textbf{Operation} \\ \hline
        Word Embedding           & $c \in \mathbb{R}^{B \times C}$ & $E(c) \in \mathbb{R}^{B \times C \times d}$ & $E(c) = \text{Word\_Embedding}(c)$ \\ \hline
        Flatten             & $E(c) \in \mathbb{R}^{B \times C \times d}$ & $F(E(c)) \in \mathbb{R}^{B \times (C \times d)}$ & $F(E(c)) = \text{Flatten}(E(c))$ \\ \hline
        Linear  & $F(E(c)) \in \mathbb{R}^{B \times (C \times d)}$ & $L_1(F(E(c))) \in \mathbb{R}^{B \times 32}$ & $L_1 = \text{PReLU}(\text{Linear}(F(E(c))))$ \\ \hline
        Linear & $L_1 \in \mathbb{R}^{B \times 32}$ & $L_2(L_1) \in \mathbb{R}^{B \times 16}$ & $L_2 = \text{PReLU}(\text{Linear}(L_1))$ \\ \hline
        Linear & $n \in \mathbb{R}^{B \times N}$ & $L_n(n) \in \mathbb{R}^{B \times 16}$ & $L_n = \text{PReLU}(\text{Linear}(n))$ \\ \hline
        Addition            & $L_2 + L_n \in \mathbb{R}^{B \times 16}$ & $S \in \mathbb{R}^{B \times 16}$ & $S = \text{PReLU}(L_2 + L_n)$ \\ \hline
        Linear (classify)   & $S \in \mathbb{R}^{B \times 16}$ & $y \in \mathbb{R}^{B \times K}$ & $y = \text{Linear}(S)$ \\ \hline
    \end{tabular}
    }
    \caption{EF-Net Architecture Details}
    \label{tab:model_architecture}
\end{table*}

%Fig \ref{fig:efnet_arch} shows the illustration of our proposed architecture.
%Our model processes numerical/categorical columns separately, followed by feature fusion. Numerical inputs are passed through a Linear layer and PReLU activation, while word embeddings of categorical column undergo Flattening, Linear layers, and PReLU. The fused features are processed through a PReLU layer and classified into various outcomes such as "Home," "Admitted," "Transfer," and others.

The model was trained using cross-entropy loss:
\[
\mathcal{L} = -\frac{1}{B} \sum_{i=1}^{B} \sum_{j=1}^{K} y_{ij}^{\text{true}} \log(\hat{y}_{ij}),
\]
where \( \hat{y}_{ij} = \frac{e^{y_{ij}}}{\sum_{l=1}^{K} e^{y_{il}}} \) is the softmax output, and \( y_{ij}^{\text{true}} \) is the true label.

\subsection{Ensemble of XGBoost and EF-Net using Soft-Voting}
In order to generate more accurate predictions, ensemble learning is a machine learning technique that combines two or more learners. Our ensemble model used EF-Net and XGBoost. Since our dataset is tabular, XGBoost is more effective when dealing with tabular data. 
The weighted average of predicted probabilities is used to make the final prediction.
% Additionally, we incorporate XGBoost into our EF-Net model to achieve superior results in the ensemble voting technique. 
Soft voting was utilized for ensemble voting. Soft voting is an algorithm that can combine the predictions of multiple classifiers using the probability of predictions. We use soft voting instead of hard voting because it is initially more precise than hard voting. Secondly, it is more resistant to data disturbance. Thirdly, flexible voting is less likely to make errors when the individual classifiers are not very confident in their predictions.
\[
\hat{y} = \frac{1}{M} \sum_{m=1}^{M} \hat{y}_m,
\]
 where \( M \) represents the number of models in the ensemble and $\hat{y}_m$ denotes the prediction of $m$-th model.

\section{Result Analysis}

%\subsection{Experimental results}
The performance of the selected models is thoroughly examined in this section, with a focus on precision, recall, F1-score, accuracy, and AUC.
% From stratified split we got a training set of 629,145 samples, with 14 numerical and 77 categorical features, while the validation and test sets each had 209,715 samples. 
In our EF-Net models, we implemented certain hyperparameters such as an optimizer, batch size, learning rate, and epochs.As shown in \Cref{tab:hyperparameters} The Adam optimizer is a fast-converging, efficient tool that employs adaptive learning rates, bias correction, and minimal hyperparameter tuning to address a diverse array of tasks. We employed Max Depth of 8 and Max Leaves of 100 in XGBoost. 

\begin{table}[ht]
\centering
\resizebox{0.7\linewidth}{!}
{\begin{tabular}{llr}
\toprule
\textbf{Hyperparameter} & \textbf{Model} & \textbf{Value} \\
\midrule
Learning Rate & EF-Net & 0.001 \\
Batch Size & EF-Net & 64 \\
Epochs & EF-Net & 50 \\
Optimizer & EF-Net & Adam \\
Max Depth & XGBoost & 8 \\
Max Leaves & XGBoost & 100 \\
\bottomrule
\end{tabular}}
\caption{Hyperparameters used in EF-Net and XGBoost.}
\label{tab:hyperparameters}
\end{table}

The performance metrics of various approaches on test data are presented in \Cref{table_3}. We analyse MIMIC-IV-ED v2.0 dataset\cite{mimic-iv-ed} in four distinct stages for predicting multiclass ER disposition. 
%We evaluate our proposed models using performance metrics such as accuracy, precision, recall, F1 score and AUC. 
%Begin with employing the neural network baseline model.
As shown in \Cref{table_3} the accuracy of our EF-Net model was 95.33\%, surpassing both the neural network baseline and XGBoost. In addition, we constructed ensemble models with EF-Net and XGBoost, achieving the greatest accuracy of 96\%. Additionally, we conducted a comparison between our model and other relevant papers. As shown in \Cref{table_4} our proposed approach outperform  than \cite{kuo2024ensemble} for forecasting ER disposition on \cite{mimic-iv-ed} dataset. 
 
\begin{table}[htbp]
\centering
\resizebox{\linewidth}{!}
{\begin{tabular}{|c|c|c|c|c|c|}
\hline
\textbf{Model}                     & \textbf{Accuracy} & \textbf{Precision} & \textbf{Recall} & \textbf{F1-Score} &\textbf{AUC} \\ \hline
\begin{tabular}[c]{@{}c@{}}Neural\\ Network\\ (Baseline)\end{tabular} & 83.19                  & 83    & 83    & 82    & 92     \\ \hline
XGBoost                   & 84       & 84        & 84     & 82       & 96    \\ \hline
EF-Net                    & 95.33    & 95.18     & 95.33  & 95.21    & 98.89 \\ \hline
\begin{tabular}[c]{@{}c@{}}Ensemble\\
(Soft voting)\end{tabular}   & 96 & 95.44 & 95.55 & 95.33 & 99.04 \\ \hline
\end{tabular}}
\caption{Comparison of Different Approaches in Our Study}
\label{table_3}
\end{table}
 
\begin{table}[ht]
\centering
\renewcommand{\arraystretch}{1.2}
\resizebox{\linewidth}{!}
{%
\begin{tabular}{|c|c|c|c|}
\hline
\textbf{Method} & \textbf{Approach} & \textbf{Accuracy} & \textbf{AUC} \\ \hline
Kuo \emph{et al.} \cite{kuo2024ensemble} & Ensemble             & 93                         & 97       \\ \hline
\multirow{2}{*}{\textbf{Proposed Model}} 
& \textbf{EF-Net}          & \textbf{95.33}         & \textbf{98.89} \\ \cline{2-4}
 & \textbf{\begin{tabular}[c]{@{}c@{}}Ensemble\\ (Soft Voting)\end{tabular}}          & \textbf{96}          & \textbf{99.04}   \\ \hline
\end{tabular}%
}
\caption{Comparison with Different Study}
\label{table_4}
\end{table}

\begin{figure}[ht]
    \centering
    \begin{subfigure}[b]{1.15\linewidth}
        \centering
        \includegraphics[width=\linewidth]{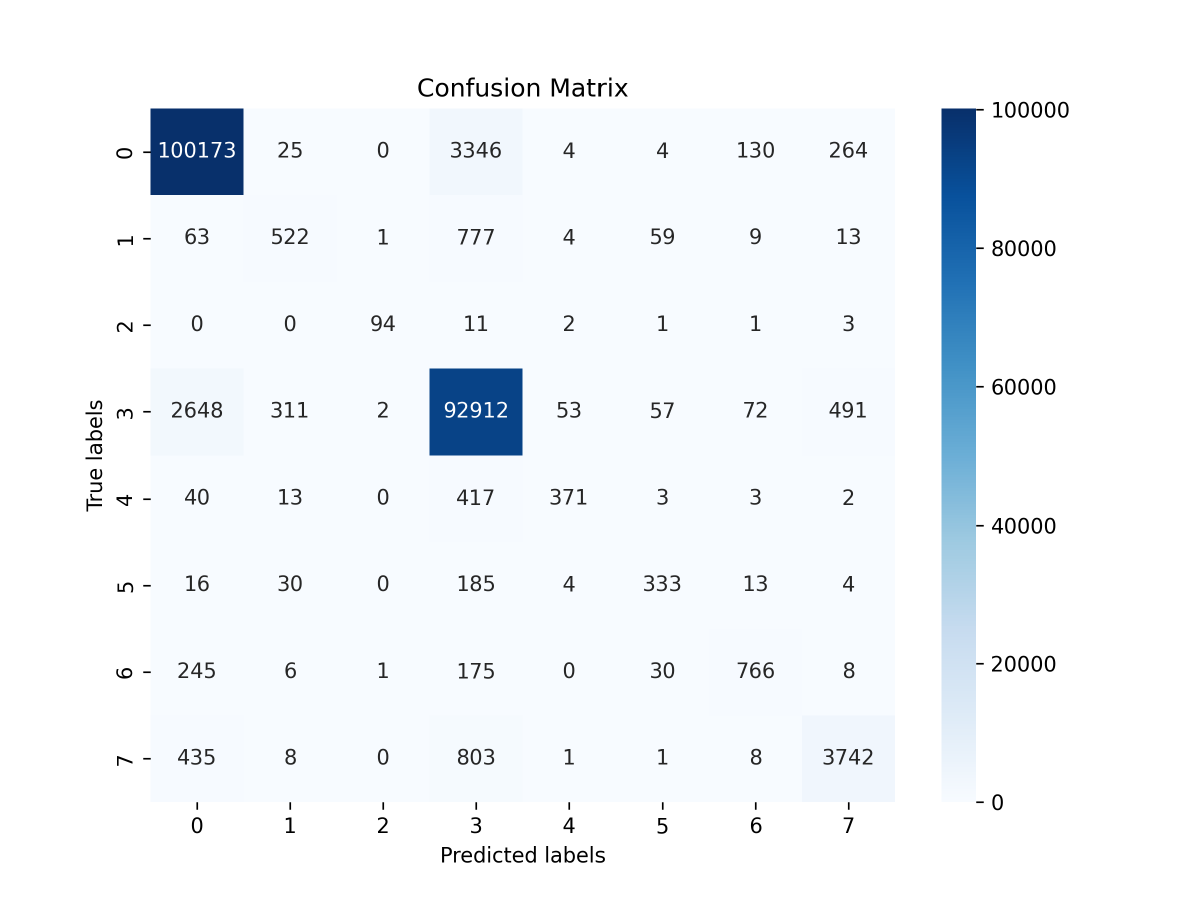}
        \caption{Confusion matrix of our proposed network, EF-Net.}
        \label{fig:conf_efnet}
    \end{subfigure}
    
    \vspace{1em} % Adds some space between the subfigures

    \begin{subfigure}[b]{1.15\linewidth}
        \centering
        \includegraphics[width=\linewidth]{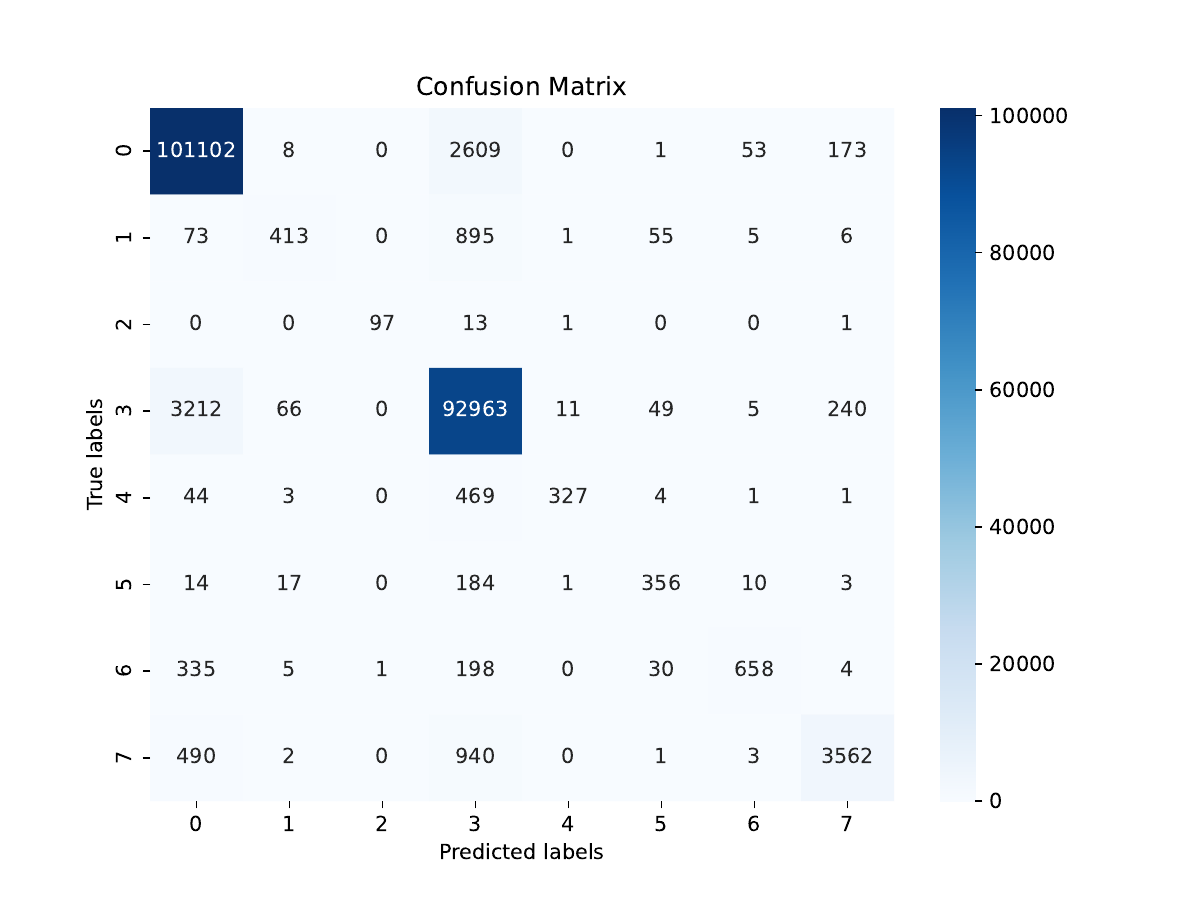}
        \caption{Confusion matrix of Ensemble Model.}
        \label{fig:conf_ensemble_model}
    \end{subfigure}
    
    \caption{Comparison of confusion matrices for EF-Net and Ensemble Model.}
    \label{fig:conf_matrices}
\end{figure}

The performance of our proposed model EF-Net, and Ensemble Model can be evaluated by comparing their respective confusion matrices, as shown in Figure \ref{fig:conf_matrices}. These matrices provide insights into the classification capabilities of both models, particularly in terms of their ability to correctly classify instances across different classes.

Data imbalance impacts the EF-Net and Ensemble models' performance on minority classes, such as 1 and 4, making them tougher to predict correctly, according to the confusion matrices for these models (Figure  \ref{fig:conf_efnet} and Figure \ref{fig:conf_ensemble_model}). With greater representation in the training data, majority classes like 0 and 3 exhibit better prediction performance.
% In Figure \ref{fig:conf_efnet} (EF-Net's confusion matrix), class 0 and class 3 show strong performance, with high true positive counts of 101,102 and 92,963, respectively, indicating the model's focus on majority classes. However, EF-Net struggles with minority classes, such as class 6 and class 7, with lower true positive counts (658 for class 6 and 3,562 for class 7) and relatively few misclassifications.
% In Figure \ref{fig:conf_ensemble_model} (Ensemble Model's confusion matrix), class 0 and class 3 similarly show excellent performance, with the same high true positive counts of 101,102 and 92,963, as in EF-Net. For minority classes, such as class 6 and class 7, the Ensemble Model offers slight improvements by reducing misclassifications, but the true positive counts (658 for class 6 and 3,562 for class 7) remain the same as EF-Net, reflecting the persistent challenge of class imbalance.

\begin{figure}[ht]
    \centering
    \includegraphics[width=1.0\linewidth]{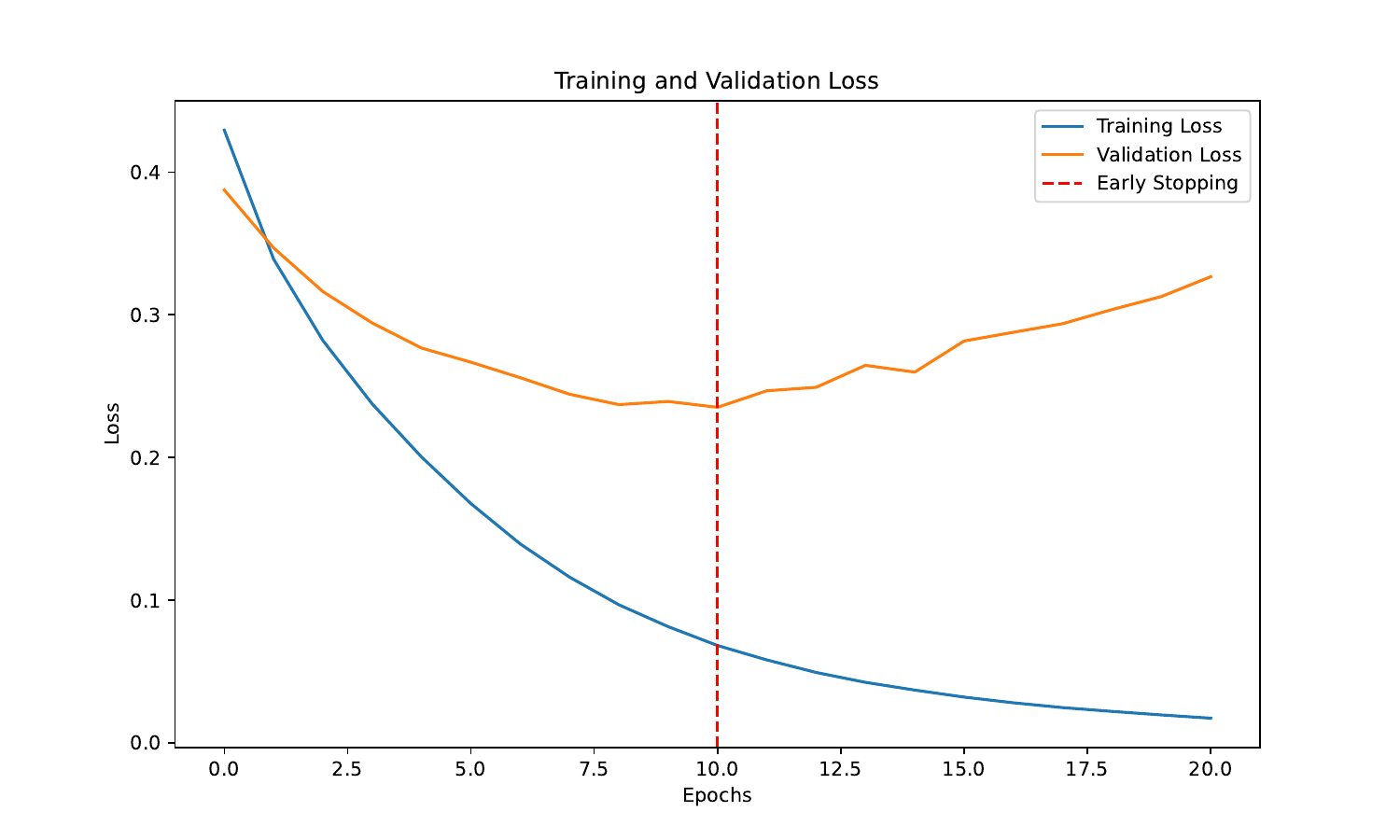}
    \caption{Training vs Validation Loss. Early stopping criteria is used to avoid overfitting of the model.}
    \label{fig:loss_plot}
\end{figure}
In order to avoid overfitting, which happens when a model performs well on training data but badly on unknown data, early stopping is a method employed during training. Following each training period, the early stopping criteria tracks the model's performance on a validation set. Training is stopped early if, after a predetermined number of successive epochs, the validation loss does not improve. This method strikes a compromise between underfitting and overfitting by ensuring that the model parameters are preserved at the point where the validation performance is at its best.
Our dataset has a class imbalance problem, which increases the likelihood of overfitting in the model. After the model began overfitting to the majority class, we used the early stopping criterion to stop it from training further which is shown in figure  \ref{fig:loss_plot}.
\\
 Figure \ref{fig:loss_plot} shows the loss of training and validation of EF-Net over 20 epochs. The training loss gradually decreases as training progresses, demonstrating how well the model fits the training set. It appears that the model starts to overfit when the validation loss initially drops but then begins to increase after some time. To mitigate this overfitting, we apply an early stopping criterion.

\section{Limitation and future work}
The study we conducted has limitations. The first limitation is that the predictive model's generalizability may be restricted by the fact that the samples were collected from a single institution. Future research can choose to gather data from an increased number of institutions in order to make more reliable improvements to the results. Secondly, this study did not include laboratory and image data, which implies that future research may incorporate these distinct types of data. Thirdly, our target variable is imbalanced. Stratify was employed to address the imbalance. We'll balance the disposition in our future work.

\section{Conclusion}
This study is developing an EF-Net and ensemble learning model to predict patient disposition in EDs to address overcrowding. Through the integration of structured and unstructured data, the model can assist clinicians in predicting discharge outcomes, thereby reducing overload.
\\Our research is designed to contribute to the development of a predictive model for the multiclass classification of dispositions in emergency units, which is based on both numerical and categorical features. In the EF-Net model, word embedding is vital. As a comparison, the baseline neural network demonstrated inferior accuracy in the absence of word embedding. The patient's significance can be prioritized by identifying the disposition classification.

\bibliographystyle{ieeetr}
\bibliography{references.bib}

% \section*{References}

% \begin{thebibliography}{00}
% \end{thebibliography}
% \vspace{12pt}

\end{document}